\documentclass{article}
\usepackage{latexsym}
\usepackage{aaai}
\usepackage{amssymb}
\usepackage{epsfig}

\newtheorem{DEF}{Definition}[section]
\newtheorem{EX}[DEF]{Example}
\newcommand{\qed}{\hspace*{\fill}$\Box$}
\newenvironment{Definition}{\begin{DEF}\rm}{\qed\end{DEF}}
\newenvironment{example}{\begin{EX}\rm}{\qed\end{EX}}

\begin{document}
\title{On the semantics of merging}
\author{
Thomas Meyer \\
Department of Computer Science\\
School of Information Technology \\
University of Pretoria\\
Pretoria, 0002, South Africa\\
e-mail: tmeyer@cs.up.ac.za}
\date{}
\maketitle
\begin{abstract}
Intelligent agents are often faced with the problem of trying to merge possibly conflicting
pieces of information obtained from different sources into a consistent view of the world. 
We propose a framework for the modelling of such merging operations with roots in 
the work of Spohn \cite{Spohn:88a,Spohn:91a}. 
Unlike most approaches we focus on the merging of epistemic states, not 
knowledge bases.  
We construct a number of plausible merging operations and measure them against
various properties that merging operations ought to satisfy. Finally, we discuss the connection
between merging and the use of infobases \cite{Meyer:99c,Meyer-ea:2000a}.
\end{abstract}
\section{Introduction}
To be able to operate in its environment it is necessary for an intelligent agent to have a consistent 
view of the world. This demand is often complicated by the fact that such agents receive conflicting
pieces of information from different sources. The process of combining possibly inconsistent pieces of 
information, known as \emph{merging}, has many applications and has started to receive more attention
recently \cite{Borgida-ea:84a,Lin:96a,Baral-ea:91a,Baral-ea:92a,Konieczny-ea:98a,Liberatore-ea:98a,Revesz:93a,Revesz:97a,Subrahmanian:94a}. 
In this paper we propose a framework for 
the modelling of merging operations. The proposal has its roots 
in the work of Spohn \cite{Spohn:88a,Spohn:91a}. Unlike most approaches we adopt a 
description of merging on  the level of \emph{epistemic states} instead of  
\emph{knowledge bases}.

First we give a brief introduction to 
the merging of knowledge bases, focussing on the work of Konieczny and Pino-P{\'e}rez
\cite{Konieczny-ea:98a}. This is followed by a  
description of our framework for the merging of epistemic states. 
Then we 
construct a number of merging operations and show how they measure up to proposed properties
of merging. Finally, we discuss links between merging and the 
\emph{infobases} of Meyer \cite{Meyer:99c}. 

We assume a finitely generated
propositional language $L$ closed under the usual propositional connectives, and with a classical
model-theoretic semantics. $U$ is the set of interpretations of $L$ and $M(\alpha)$ is
the set of models of $\alpha\in L$. Classical entailment 
is denoted by $\vDash$. 
We use $\sqcup$ to denote the concatenation of lists.
We let $x^{n}$ denote the list
consisting of $n$ versions of $x$. The length of a list $l$ is denoted by $\left|l\right|$. 
\section{Merging knowledge bases}
In the spirit of the work of Katsuno and Mendelzon
\cite{Katsuno-ea:91a}, approaches to the merging of knowledge bases usually represent 
the beliefs of an agent as a single wff $\phi$ of $L$, known as a \emph{knowledge base}, where 
$\phi$ represents the set of all wffs entailed by $\phi$.
The goal is to construct, from a finite list of such knowledge bases, 
an appropriate consistent knowledge base in some rational fashion.
Konieczny and Pino-P{\'e}rez \cite{Konieczny-ea:98a} have proposed a general framework for
the merging of knowledge bases. A 
\emph{knowledge list} $e$ is a finite list of consistent knowledge 
bases $[\phi_{1},\ldots,\phi_{\left|e\right|}]$. Two knowledge lists $e_{1}$ 
and $e_{2}$ are \emph{element-equivalent}, written as $e_{1}\approx e_{2}$, iff 
for every element $\phi_{1}$ of $e_{1}$ there is a unique element $\phi_{2}$ (position-wise) of 
$e_{2}$ such that $\phi_{1}\equiv\phi_{2}$ and 
for every element $\phi_{2}$ of $e_{2}$ there is a unique element $\phi_{1}$ (position-wise) of 
$e_{1}$ such that $\phi_{2}\equiv\phi_{1}$.
A \emph{KP-merging operation} $\delta$ is a 
function from the set of all knowledge lists to the set of all knowledge bases satisfying the 
following postulates (the KP-postulates): 
\begin{description}
\item[(KP1)] $\delta(e)\nvDash\bot$
\item[(KP2)] If $\bigwedge_{i=1}^{\left|e\right|}\phi_{i}\nvDash\bot$ then 
             $\delta(e)=\bigwedge_{i=1}^{\left|e\right|}\phi_{i}$
\item[(KP3)] If $e_{1}\approx e_{2}$ then $\delta(e_{1})\equiv\delta(e_{2})$
\item[(KP4)] If $\phi_{1}\wedge \phi_{2}\vDash\bot$ then $\delta([\phi_{1}]\sqcup[\phi_{2}])\nvDash \phi_{1}$
\item[(KP5)] $\delta(e_{1})\wedge\delta(e_{2})\vDash\delta(e_{1}\sqcup e_{2})$
\item[(KP6)] If $\delta(e_{1})\wedge\delta(e_{2})\nvDash\bot$ then 
             $\delta(e_{1}\sqcup e_{2})\vDash\delta(e_{1})\wedge\delta(e_{2})$
\end{description}
Konieczny and Pino-P{\'e}rez also distinguish between two subclasses of merging operations. 
An \emph{arbitration} operation tries to take as many differing opinions as possible into account,  
while the intuition associated with 
\emph{majority} operations is that the opinion of the majority should prevail.
They initially propose the following postulates for arbitration and majority operations.
\begin{description}
\item[(arb)] $\forall n\ \delta(e\sqcup{\phi}^n)=\delta(e\sqcup[\phi])$
\item[(maj)] $\exists n\ \delta(e\sqcup \phi^{n})\vDash\phi$
\end{description}
It turns out that there is no KP-merging operation satisfying (arb). Unlike Konieczny and
Pino-P{\'e}rez we are of the opinion that it is not (arb) that is at fault, but some of the 
KP-postulates. Below we argue against the inclusion of (KP4) and (KP6) as postulates
that need to be satisfied by all merging operations.

\section{Merging epistemic states}
In this section we discuss merging on the level of epistemic states.
We see an epistemic state as providing a
plausibility ranking of the interpretations of $L$;
the lower the number assigned to an interpretation, the more plausible it is deemed to be. 
\begin{Definition}
An epistemic state $\Phi$ is a function from $U$ to the set of natural 
numbers. Given an epistemic state $\Phi$, the knowledge base associated with $\Phi$, denoted by 
$\phi_{\Phi}$, is some $\alpha\in L$ such that $M(\alpha)=\{u\mid \Phi(u)=0\}$. 
\end{Definition}
This representation of an epistemic state and its associated knowledge base can be traced back 
to the work of Spohn \cite{Spohn:88a,Spohn:91a}. It should be clear that an epistemic state 
with an inconsistent associated knowledge base still contains useful information.

An \emph{epistemic list} $E= [\Phi^{E}_{1},\ldots,\Phi^{E}_{\left|E\right|}]$ is a finite 
list of epistemic states.
\begin{figure}
\begin{center}
\begin{picture}(0,0)%
\epsfig{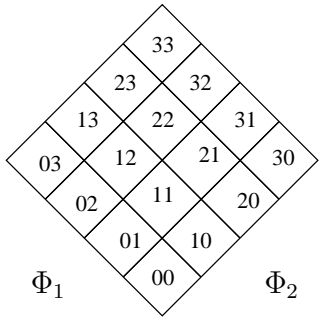}%
\end{picture}%
\setlength{\unitlength}{4144sp}%
\begingroup\makeatletter\ifx\SetFigFont\undefined%
\gdef\SetFigFont#1#2#3#4#5{%
  \reset@font\fontsize{#1}{#2pt}%
  \fontfamily{#3}\fontseries{#4}\fontshape{#5}%
  \selectfont}%
\fi\endgroup%
\begin{picture}(1888,1887)(1099,-1697)
\put(1261,-1546){\makebox(0,0)[lb]{\smash{\SetFigFont{12}{14.4}{\rmdefault}{\mddefault}{\updefault}$\Phi_{1}$}}}
\put(2656,-1546){\makebox(0,0)[lb]{\smash{\SetFigFont{12}{14.4}{\rmdefault}{\mddefault}{\updefault}$\Phi_{2}$}}}
\end{picture}
\end{center}
\caption{A pictorial representation of an epistemic list containing two epistemic states 
$\Phi_{1}$ and $\Phi_{2}$.
The sequence of two digits in each cell above indicates the natural numbers associated with 
interpretations by the two epistemic states. A cell containing the sequence $ij$ indicates the 
placement of those interpretations assigned the value $i$ by $\Phi_{1}$ and assigned the value
$j$ by $\Phi_{2}$.}
\label{Fig:Basic}
\end{figure}
It is instructive to view an epistemic list pictorially as in figure
\ref{Fig:Basic}. While such a pictorial view is only useful in representing epistemic 
lists containing two elements, it serves as a good foundation for understanding the principles 
underlying the merging of epistemic states in general.

For any epistemic state $\Phi$, let 
\[\min(\Phi)=\min\{\Phi(u)\mid u\in U\}
\]
let
\[
\max(\Phi)=\max\{\Phi(u)\mid u\in U\}
\] 
and for an epistemic list $E$, let 
\[
\max(E)=\max\{\max(\Phi^{E}_{i})\mid 1\leq i\leq\left|E\right|\}.
\]
For an epistemic list $E$ and $u\in U$ we let $\min^{E}(u)$ be equal to 
\[
\min\{\Phi^{E}_{i}(u)\mid 1\leq i\leq\left|E\right|\}
\]
and we let $\max^{E}(u)$ be equal to 
\[
\max\{\Phi^{E}_{i}(u)\mid 1\leq i\leq\left|E\right|\}.
\] 
We denote by $seq(E)$ the set of all sequences of length $\left|E\right|$ of natural numbers, 
ranging from $0$ to $\max(E)$.
We denote by $seq_{\leq}(E)$ the subset of $seq(E)$ of all sequences that are ordered 
non-decreasingly, and by $seq_{\geq}(E)$ the subset of $seq(E)$ of all sequences that are ordered 
non-increasingly. For $u\in U$, we let $s^{E}(u)$ be the sequence containing the 
natural numbers $\Phi^{E}_{1}(u),\ldots,\Phi^{E}_{\left|E\right|}(u)$ in that order, we let 
$s^{E}_{\leq}(u)$ be the sequence $s^{E}(u)$ ordered non-decreasingly,  
and we let $s^{E}_{\geq}(u)$ be the sequence $s^{E}(u)$ ordered non-increasingly. Clearly 
$s^{E}(u)\in seq(E)$, $s^{E}_{\leq}(u)\in seq_{\leq}(E)$ and $s^{E}_{\geq}(u)\in seq_{\geq}(E)$. 
Given any set $seq$ of finite sequences of natural numbers and a total preorder $\sqsubseteq$ on 
$seq$, we define the function 
$\Omega^{seq}_{\sqsubseteq}:seq\rightarrow \{0,\ldots,\left|seq\right|-1\}$
by assigning natural numbers to the elements of 
$seq$ in the order imposed by $\sqsubseteq$, starting by assigning $0$ to the elements 
lowest down in $\sqsubseteq$. We denote the 
\emph{lexicographic} ordering on $seq$ by $\sqsubseteq_{lex}$. 

A \emph{merging operation on epistemic states} $\Delta$ is a function from the set of all
non-empty epistemic lists to the set of all epistemic states. 
We propose the following basic properties for the merging of epistemic states:
\begin{description}
\item[(E1)] $\exists u$ s.t. $\Delta(E)(u)=0$
\item[(E2)] If $\Phi^{E}_{i}(u)=\Phi^{E}_{j}(u)\ \forall i,j$ such that $1\leq i,j\leq\left|E\right|$ and 
            $s^{E}_{\leq}(u)\sqsubset_{lex}s^{E}_{\leq}(v)$ then 
	    $\Delta(E)(u)<\Delta(E)(v)$ 
\item[(E3)] If $\Phi^{E}_{i}(u)\leq\Phi^{E}_{i}(v)\ \forall i$ such that $1\leq i\leq\left|E\right|$ then 
            $\Delta(E)(u)\leq\Delta(E)(v)$
\item[(E4)] If $\Delta(E)(u)\leq\Delta(E)(v)$ then $\Phi^{E}_{i}(u)\leq\Phi^{E}_{i}(v)$ for 
            some $i$ such that $1\leq i\leq\left|E\right|$ 
\end{description}
(E1) is a restatement of (KP1) and (E2) generalises (KP2). (E3) states that if 
all epistemic states in $E$ agree that $u$ is at least as plausible as $v$, then so should the resulting 
epistemic state. (E4) expects justification 
for regarding an interpretation $u$ as at least as plausible as $v$: there has to be at least one 
epistemic state in $E$ which regards $u$ as at least as plausible as $v$.
The following fundamental principle for the merging of epistemic states follows easily from (E3):
\begin{description}
\item[(Unit)]  If $\Phi^{E}_{i}(u)=\Phi^{E}_{i}(v)\ \forall i$ such that $1\leq i\leq\left|E\right|$ then 
            $\Delta(E)(u)=\Delta(E)(v)$
\end{description}
(Unit) requires interpretations that are treated identically by all epistemic states in an epistemic
list to be treated identically in the epistemic state resulting from a merging operation. 

Two epistemic lists $E_{1}$ 
and $E_{2}$ are \emph{element-equivalent}, written as $E_{1}\approx E_{2}$, iff for 
every element $\Phi_{1}$ of $E_{1}$ there is a unique element $\Phi_{2}$ (position-wise) of $E_{2}$
such that $\Phi_{1}=\Phi_{2}$ and for every 
element $\Phi_{2}$ of $E_{2}$ there is a unique element $\Phi_{1}$ (position-wise) of $E_{1}$
such that $\Phi_{2}=\Phi_{1}$. The following property is a generalisation of (KP3). 
It requires merging to be commutative.
\begin{description}
\item[(Comm)] $E_{1}\approx E_{2}$ implies $\Delta(E_{1})=\Delta(E_{2})$
\end{description}
We do not think that (Comm) should hold for all merging operations. Instead, (Comm) should be seen 
as a postulate picking out an interesting subclass of merging operations.

For a finite list of epistemic lists
$\mathcal{E}=[E_{1},\ldots,E_{\left|\mathcal{E}\right|}]$, let $\Delta(\mathcal{E})$ denote the
epistemic list $[\Delta(E_{1}),\ldots,\Delta(E_{\left|\mathcal{E}\right|})]$.
We consider the following properties:
\begin{description}
\item[(E5)] If $\Delta(E_{i})(u)\leq\Delta(E_{i})(v)\ \forall i$ such that 
$1\leq i\leq\left|\mathcal{E}\right|$ then 
            $\Delta(\bigsqcup_{i=1}^{\left|\mathcal{E}\right|}E_{i})(u)\leq
	     \Delta(\bigsqcup_{i=1}^{\left|\mathcal{E}\right|}E_{i})(v)$
\item[(E6)] If $\Delta(\bigsqcup_{i=1}^{\left|\mathcal{E}\right|}E_{i})(u)\leq
             \Delta(\bigsqcup_{i=1}^{\left|\mathcal{E}\right|}E_{i})(v)$
            then for some $i$ such that $1\leq i\leq\left|\mathcal{E}\right|$, 
	    $\Delta(E_{i})(u)\leq\Delta(E_{i})(v)$ for some $1\leq i\leq\left|\mathcal{E}\right|$ 
\end{description}
(E5) generalises (E3) and (E6) generalises (E4). In fact, (E5) also implies (KP5).

The arbitration postulate (arb) and the majority postulate (maj) can be generalised as follows:
\begin{description}
\item[(Arb)] $\forall n\ \Delta(E\sqcup[\Phi])(u)=\Delta(E\sqcup\Phi^{n})(u)$
\item[(Maj)] $\exists n$ s.t. $\forall u,v\in U,\ \Phi(u)\leq\Phi(v)$ if 
             $\Delta(E\sqcup\Phi^{n})(u)\leq\Delta(E\sqcup\Phi^{n})(v)$
\end{description}
We have not provided a generalised version of (KP4). The reason is that we do not regard it as a suitable 
postulate for merging. Our basic argument is that the models of a knowledge base 
associated with an epistemic state $\Phi_{1}$ may sometimes be given such an implausible ranking
by an epistemic state $\Phi_{2}$ that it would seem reasonable to exclude all these models from the 
models of $\phi_{\Delta([\Phi_{1}]\sqcup[\Phi_{2}])}$. It is worthwhile noting that
none of the merging operations we consider below satisfies (KP4). 
Similarly, we have not provided a generalised version of (KP6) since we regard it as too strong a 
condition to impose on all merging operations.%
\footnote{(E6) can be regarded as a generalised version of a weaker form of (KP6), but (KP6) does not 
follow from (E6).} 
Below we shall encounter a number of reasonable merging operations which do not satisy 
(KP6).
\begin{figure}[t!]
\begin{center}
\begin{picture}(0,0)%
\epsfig{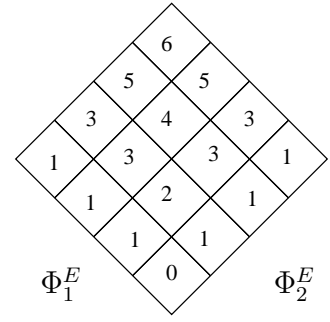}%
\end{picture}%
\setlength{\unitlength}{4144sp}%
\begingroup\makeatletter\ifx\SetFigFont\undefined%
\gdef\SetFigFont#1#2#3#4#5{%
  \reset@font\fontsize{#1}{#2pt}%
  \fontfamily{#3}\fontseries{#4}\fontshape{#5}%
  \selectfont}%
\fi\endgroup%
\begin{picture}(1888,1887)(1129,-1832)
\put(1291,-1681){\makebox(0,0)[lb]{\smash{\SetFigFont{12}{14.4}{\rmdefault}{\mddefault}{\updefault}$\Phi^{E}_{1}$}}}
\put(2686,-1681){\makebox(0,0)[lb]{\smash{\SetFigFont{12}{14.4}{\rmdefault}{\mddefault}{\updefault}$\Phi^{E}_{2}$}}}
\end{picture}
\end{center}
\caption{A representation of the merging operation $\Delta_{ls}$. 
The number in a cell represents the numbers that the appropriate merging operation assigns to the 
interpretations contained in that cell before normalisation.}
\label{Fig:ls1}
\end{figure}
\section{Constructing merging operations}
Konieczny and Pino-P{\'e}rez \cite{Konieczny-ea:98a} discuss several merging operations on knowledge 
bases using Dalal's measure of distance between interpretations \cite{Dalal:88a}. For any two 
interpretations $u$ and $v$, let $dist(u,v)$ denote the number of propositional atoms on which $u$ and $v$ 
differ. The distance $Dist(\phi,u)$ between a knowledge base $\phi$ and an interpretation $u$ is defined 
as follows: $Dist(\phi,u) = \min\{dist(u,v)\mid v\in M(\phi)\}$. It is clear that this distance measure 
can be used to
define an epistemic state $\Phi$ as follows: 
\[
\forall u\in U,\ \Phi(u)=Dist(\phi,u).
\]
It is easily 
seen that
$\Phi(u)=0$ iff $u\in M(\phi)$ and therefore $\phi_{\Phi}\equiv \phi$. Many of the merging operations
on epistemic states that we propose below are appropriate generalisations of these
merging operations on knowledge bases.

When reading through the remainder of this section, the reader should observe that 
the construction of every merging operation consists of two 
steps. In the first step natural numbers are assigned to interpretations. After 
the completion of this step it will often be the case that \emph{none} of the interpretations have been 
assigned the value $0$. To ensure compliance with (E1) the second step performs 
an appropriate uniform subtraction of values which we shall refer to as \emph{normalisation}.

\subsection{Arbitration}
Inspired by an arbitration operation proposed by 
Liberatore and Schaerf \cite{Liberatore-ea:98a} we propose the following 
two merging operations on epistemic states.
\begin{Definition}
\begin{enumerate}
\item Let $\Phi^{E}_{ls}(u)=2\min^{E}(u)$ if 
$\Phi^{E}_{i}(u)=\Phi^{E}_{j}(u)$ for $1\leq i,j\leq\left|E\right|$, and 
$\Phi^{E}_{ls}(u)=2\min^{E}(u)+1$ otherwise. Then 
$\Delta_{ls}(E)(u)=\Phi^{E}_{ls}(u) - \min(\Phi^{E}_{ls})$.
\item Let 
$\Phi^{E}_{Rls}(u)=\Omega^{seq_{\leq}(E)}_{\sqsubseteq_{lex}}(s^{E}_{\leq}(u))$.  
Then
$\Delta_{Rls}(E)(u)=\Phi^{E}_{Rls}(u) - \min(\Phi^{E}_{Rls})$.
\end{enumerate}
\end{Definition}
\begin{figure}[t]
\begin{center}
\begin{picture}(0,0)%
\epsfig{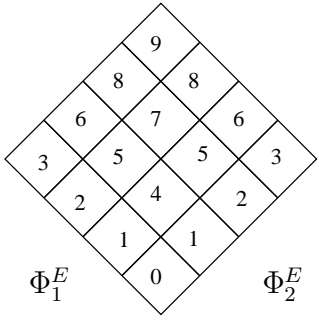}%
\end{picture}%
\setlength{\unitlength}{4144sp}%
\begingroup\makeatletter\ifx\SetFigFont\undefined%
\gdef\SetFigFont#1#2#3#4#5{%
  \reset@font\fontsize{#1}{#2pt}%
  \fontfamily{#3}\fontseries{#4}\fontshape{#5}%
  \selectfont}%
\fi\endgroup%
\begin{picture}(1888,1887)(4564,-1832)
\put(4726,-1681){\makebox(0,0)[lb]{\smash{\SetFigFont{12}{14.4}{\rmdefault}{\mddefault}{\updefault}$\Phi^{E}_{1}$}}}
\put(6121,-1681){\makebox(0,0)[lb]{\smash{\SetFigFont{12}{14.4}{\rmdefault}{\mddefault}{\updefault}$\Phi^{E}_{2}$}}}
\end{picture}
\end{center}
\caption{A representation of the merging operation $\Delta_{Rls}$ 
The number in a cell represents the numbers that the appropriate merging operation assigns to the 
interpretations contained in that cell before normalisation.}
\label{Fig:ls2}
\end{figure}
Figure \ref{Fig:ls1} contains a pictorial representation of $\Delta_{ls}$ and figure \ref{Fig:ls2}
a pictorial representation of $\Delta_{Rls}$.
It can easily be shown that $\Delta_{Rls}$ is a refined version of $\Delta_{ls}$.
Both satisfy (E1)-(E6) and (Comm), neither satisfies (Maj), and only $\Delta_{Rls}$ satisfies (KP6).
Moreover, $\Delta_{ls}$ satisfies (Arb) but $\Delta_{Rls}$ does not. So, while both are 
valid merging operations, $\Delta_{Rls}$ should not be seen as an arbitation operation.

Next we consider two merging operations that are generalisations of the $\delta_{\max}$ 
and $\delta_{Gmax}$ operations of Konieczny and Pino-P{\'e}rez. The former was inspired by an example of 
Revesz's model-fitting operations \cite{Revesz:97a}.
\begin{figure}[b!]
\begin{center}
\begin{picture}(0,0)%
\epsfig{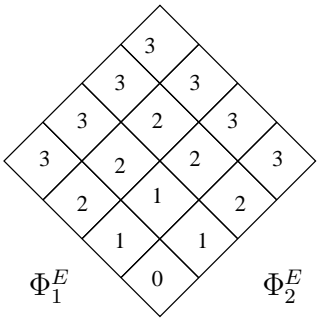}%
\end{picture}%
\setlength{\unitlength}{4144sp}%
\begingroup\makeatletter\ifx\SetFigFont\undefined%
\gdef\SetFigFont#1#2#3#4#5{%
  \reset@font\fontsize{#1}{#2pt}%
  \fontfamily{#3}\fontseries{#4}\fontshape{#5}%
  \selectfont}%
\fi\endgroup%
\begin{picture}(1888,1887)(1129,-1832)
\put(1291,-1681){\makebox(0,0)[lb]{\smash{\SetFigFont{12}{14.4}{\rmdefault}{\mddefault}{\updefault}$\Phi^{E}_{1}$}}}
\put(2686,-1681){\makebox(0,0)[lb]{\smash{\SetFigFont{12}{14.4}{\rmdefault}{\mddefault}{\updefault}$\Phi^{E}_{2}$}}}
\end{picture}
\end{center}
\caption{A representation of the merging operation $\Delta_{\max}$.
The number in a cell represents the numbers that the 
appropriate merging operation assigns to the interpretations contained in that cell before 
normalisation.}
\label{Fig:minmax1}
\end{figure} 
\begin{Definition}
\begin{enumerate}
\item Let $\Phi^{E}_{\max}(u)=\max^{E}(u)$. Then 
$\Delta_{\max}(E)(u)=\Phi^{E}_{\max}(u) - \min(\Phi^{E}_{\max})$.
\item Let 
$\Phi^{E}_{Gmax}(u)=\Omega^{seq_{\geq}(E)}_{\sqsubseteq_{lex}}(s^{E}_{\geq}(u))$. 
Then $\Delta_{Gmax}(E)(u)=\Phi^{E}_{Gmax}(u) - \min(\Phi^{E}_{Gmax})$.
\end{enumerate}
\end{Definition}
\begin{figure}[t!]
\begin{center}
\begin{picture}(0,0)%
\epsfig{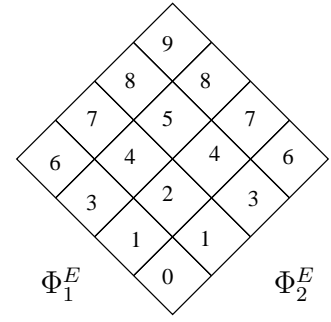}%
\end{picture}%
\setlength{\unitlength}{4144sp}%
\begingroup\makeatletter\ifx\SetFigFont\undefined%
\gdef\SetFigFont#1#2#3#4#5{%
  \reset@font\fontsize{#1}{#2pt}%
  \fontfamily{#3}\fontseries{#4}\fontshape{#5}%
  \selectfont}%
\fi\endgroup%
\begin{picture}(1888,1887)(4564,-1832)
\put(4726,-1681){\makebox(0,0)[lb]{\smash{\SetFigFont{12}{14.4}{\rmdefault}{\mddefault}{\updefault}$\Phi^{E}_{1}$}}}
\put(6121,-1681){\makebox(0,0)[lb]{\smash{\SetFigFont{12}{14.4}{\rmdefault}{\mddefault}{\updefault}$\Phi^{E}_{2}$}}}
\end{picture}
\end{center}
\caption{A representation of the merging operation $\Delta_{Gmax}$. 
The number in a cell represents the numbers that the 
appropriate merging operation assigns to the interpretations contained in that cell before 
normalisation.}
\label{Fig:minmax2}
\end{figure}  
Figure \ref{Fig:minmax1} contains a pictorial representation of $\Delta_{\max}$ and 
figure \ref{Fig:minmax2} a pictorial representation of $\Delta_{Gmax}$. 
Both satisfy (E1)-(E6), neither satisfies (Maj), and only $\Delta_{Gmax}$ satisfies (KP6).
Moreover, $\Delta_{\max}$ satisfies (Arb), but $\Delta_{Gmax}$ does not. So, analogous to the case
above, both are valid merging operations but $\Delta_{Gmax}$ should not be seen as an arbitation 
operation. The fact that we do not regard $\Delta_{Gmax}$ as an arbitration operation
is in conflict with the view of Konieczny and 
Pino-P{\'e}rez who regard $\delta_{Gmax}$ as an arbitration operation on knowledge bases even though 
it does not satisfy (arb). Conversely, Konieczny and Pino-P{\'e}rez 
do not regard $\delta_{\max}$ as a merging operation on knowledge bases since
it fails to satisfy (KP6). But we regard it as a valid arbitration
operation since it satisfies the postulates (E1)-(E6), (Comm) and (Arb).

\subsection{Consensus}
In this section we consider the idea of a \emph{consensus} operation, where agreement on the
ranking of interprerations, instead of the ranking itself, is of overriding importance.
\begin{Definition}
For $s\in seq(E)$, let  
\[
d^{E}(s) = \sum_{i=1}^{\left|E\right|}\sum_{j=i+1}^{\left|E\right|}\left|s_{i} - s_{j}\right|
\]
where $s_{i}$ denotes the $i$th element of $s$.
\begin{enumerate}
\item Define the total preorder $\sqsubseteq$ on $seq(E)$ as follows: 
$s\sqsubseteq t$ iff $d^{E}(s)\leq d^{E}(t)$. 
Let $\Phi^{E}_{cons}(u)=\Omega^{seq(E)}_{\sqsubseteq}(s^{E}(u))$.
Then 
$\Delta_{cons}(E)(u)=\Phi^{E}_{cons}(u) - \min(\Phi^{E}_{cons})$.
\item Define the total preorder $\sqsubseteq$ on $seq_{\leq}(E)$ as follows:
$s\sqsubseteq t$ iff $d^{E}(s)<d^{E}(t)$ or ($d^{E}(s)=d^{E}(t)$ and 
$s\sqsubseteq_{lex}t$). Now,  
let $\Phi^{E}_{Rcons}(u)=\Omega^{seq_{\leq}(E)}_{\sqsubseteq}(s^{E}_{\leq}(u))$. Then 
$\Delta_{Rcons}(E)(u)=\Phi^{E}_{Rcons}(u) - \min(\Phi^{E}_{Rcons})$.
\end{enumerate}
\end{Definition}
\begin{figure}[t!]
\begin{center}
\begin{picture}(0,0)%
\epsfig{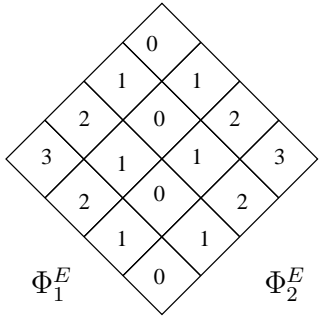}%
\end{picture}%
\setlength{\unitlength}{4144sp}%
\begingroup\makeatletter\ifx\SetFigFont\undefined%
\gdef\SetFigFont#1#2#3#4#5{%
  \reset@font\fontsize{#1}{#2pt}%
  \fontfamily{#3}\fontseries{#4}\fontshape{#5}%
  \selectfont}%
\fi\endgroup%
\begin{picture}(1888,1887)(1129,-1832)
\put(1291,-1681){\makebox(0,0)[lb]{\smash{\SetFigFont{12}{14.4}{\rmdefault}{\mddefault}{\updefault}$\Phi^{E}_{1}$}}}
\put(2686,-1681){\makebox(0,0)[lb]{\smash{\SetFigFont{12}{14.4}{\rmdefault}{\mddefault}{\updefault}$\Phi^{E}_{2}$}}}
\end{picture}
\end{center}
\caption{A representation of the merging operation $\Delta_{cons}$.
As usual, the number in a cell represents the numbers that the 
appropriate merging operation assigns to the interpretations contained in that cell before 
normalisation.}
\label{Fig:cons1}
\end{figure}
\begin{figure}[b!]
\begin{center}
\begin{picture}(0,0)%
\epsfig{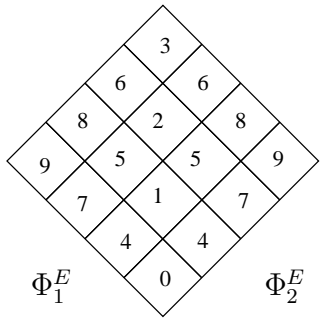}%
\end{picture}%
\setlength{\unitlength}{4144sp}%
\begingroup\makeatletter\ifx\SetFigFont\undefined%
\gdef\SetFigFont#1#2#3#4#5{%
  \reset@font\fontsize{#1}{#2pt}%
  \fontfamily{#3}\fontseries{#4}\fontshape{#5}%
  \selectfont}%
\fi\endgroup%
\begin{picture}(1888,1887)(4564,-1832)
\put(4726,-1681){\makebox(0,0)[lb]{\smash{\SetFigFont{12}{14.4}{\rmdefault}{\mddefault}{\updefault}$\Phi^{E}_{1}$}}}
\put(6121,-1681){\makebox(0,0)[lb]{\smash{\SetFigFont{12}{14.4}{\rmdefault}{\mddefault}{\updefault}$\Phi^{E}_{2}$}}}
\end{picture}
\end{center}
\caption{A representation of the merging operation $\Delta_{Rcons}$. 
As usual, the number in a cell represents the numbers that the 
appropriate merging operation assigns to the interpretations contained in that cell before 
normalisation.}
\label{Fig:cons2}
\end{figure}
Figure \ref{Fig:cons1} contains a pictorial representation of $\Delta_{cons}$ and 
figure \ref{Fig:cons2} a pictorial representation of $\Delta_{Rcons}$. 
We do not regard these two operations as suitable candidates for merging, primarily because both
fail to satisfy (E3) and (E4). Both satisfy (Unit), though.
The problem with these consensus operations seems to be that they place too strong an emphasis
on agreement and do not take the ranking of interpretations seriously enough.

\subsection{Majority}
\label{Subsec:Majority}%
We consider the following two majority operations.
\begin{Definition}
For $s\in seq(E)$, let  
\[
sum^{E}(s) = \sum_{i=1}^{\left|E\right|}s_{i}
\] 
where $s_{i}$ is the $i$th element of $s$.
\begin{enumerate}
\item Let $\Phi^{E}_{\Sigma}(u)=sum^{E}(s^{E}(u))$. Then
$\Delta_{\Sigma}(E)(u)=\Phi^{E}_{\Sigma}(u) - \min(\Phi^{E}_{\Sigma})$.
\item Define the total preorder $\sqsubseteq$ on $seq(E)$ as follows:\newline
$s\sqsubseteq t$ iff $sum^{E}(s)<sum^{E}(t)$ or \newline
($sum^{E}(s)=sum^{E}(t)$ and 
$d^{E}(s)\leq d^{E}(t)$). Now, 
let $\Phi^{E}_{R\Sigma}(u)=\Omega^{seq(E)}_{\sqsubseteq}(s^{E}(u))$. Then
$\Delta_{R\Sigma}(E)(u)=\Phi^{E}_{R\Sigma}(u) - \min(\Phi^{E}_{R\Sigma})$.
\end{enumerate}
\end{Definition}
\begin{figure}[t!]
\begin{center}
\begin{picture}(0,0)%
\epsfig{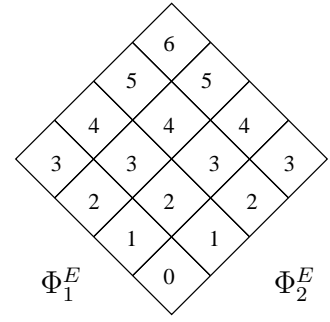}%
\end{picture}%
\setlength{\unitlength}{4144sp}%
\begingroup\makeatletter\ifx\SetFigFont\undefined%
\gdef\SetFigFont#1#2#3#4#5{%
  \reset@font\fontsize{#1}{#2pt}%
  \fontfamily{#3}\fontseries{#4}\fontshape{#5}%
  \selectfont}%
\fi\endgroup%
\begin{picture}(1888,1887)(1129,-1832)
\put(1291,-1681){\makebox(0,0)[lb]{\smash{\SetFigFont{12}{14.4}{\rmdefault}{\mddefault}{\updefault}$\Phi^{E}_{1}$}}}
\put(2686,-1681){\makebox(0,0)[lb]{\smash{\SetFigFont{12}{14.4}{\rmdefault}{\mddefault}{\updefault}$\Phi^{E}_{2}$}}}
\end{picture}
\end{center}
\caption{A representation of the merging operation $\Delta_{\Sigma}$. 
As usual, the number in a cell represents the numbers that the 
appropriate merging operation assigns to the interpretations contained in that cell before 
normalisation.}
\label{Fig:maj1}
\end{figure}
\begin{figure}[b!]
\begin{center}
\begin{picture}(0,0)%
\epsfig{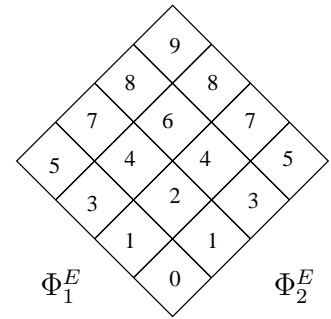}%
\end{picture}%
\setlength{\unitlength}{4144sp}%
\begingroup\makeatletter\ifx\SetFigFont\undefined%
\gdef\SetFigFont#1#2#3#4#5{%
  \reset@font\fontsize{#1}{#2pt}%
  \fontfamily{#3}\fontseries{#4}\fontshape{#5}%
  \selectfont}%
\fi\endgroup%
\begin{picture}(1888,1887)(4564,-1832)
\put(4726,-1681){\makebox(0,0)[lb]{\smash{\SetFigFont{12}{14.4}{\rmdefault}{\mddefault}{\updefault}$\Phi^{E}_{1}$}}}
\put(6121,-1681){\makebox(0,0)[lb]{\smash{\SetFigFont{12}{14.4}{\rmdefault}{\mddefault}{\updefault}$\Phi^{E}_{2}$}}}
\end{picture}
\end{center}
\caption{A representation of the merging operation $\Delta_{R\Sigma}$.
As usual, the number in a cell represents the numbers that the 
appropriate merging operation assigns to the interpretations contained in that cell before 
normalisation.}
\label{Fig:maj2}
\end{figure}
Figure \ref{Fig:maj1} contains a pictorial representation of $\Delta_{\Sigma}$ and
figure \ref{Fig:maj2} a pictorial representation of $\Delta_{R\Sigma}$.
$\Delta_{\Sigma}$ is an appropriate generalisation of an example by Lin and
Mendelzon \cite{Lin-ea:??}. It was independently proposed by Revesz \cite{Revesz:93a} as an 
example of weighted model fitting. The idea is simply to obtain the new plausibility ranking
of an interpretation by summing the plausibility rankings given by the different epistemic states. 
$\Delta_{R\Sigma}$ is $\Delta_{\Sigma}$ refined by using consensus.
Both $\Delta_{\Sigma}$ and $\Delta_{R\Sigma}$ satisfy (E1)-(E4), (Comm) and (Maj), 
and neither satisfies (Arb). But while $\Delta_{\Sigma}$ satisfies (E5)-(E6) and (KP5)-(KP6) as well, 
$\Delta_{R\Sigma}$ does not.

\subsection{Non-commutative merging}
Thus far we have restricted ourselves to the construction of \emph{commutative} merging operations --
i.e., satisfying (Comm) -- but a complete description of merging ought to take into account 
constructions such as that of Nayak \cite{Nayak:94b}, in which the merging of two epistemic states is 
obtained by a lexicographic refinement of one by the other. We present here a generalised version
of Nayak's proposal. For this case the epistemic states in an epistemic list are assumed to be
ranked according to reliability. That is, given an epistemic list
$E= [\Phi^{E}_{1},\ldots,\Phi^{E}_{\left|E\right|}]$, $\Phi^{E}_{i}$ is at least
as reliable as $\Phi^{E}_{j}$ iff $i\leq j$.
\begin{Definition}
Let $\Phi^{E}_{lex}(u)=\Omega^{seq(E)}_{\sqsubseteq_{lex}}(s^{E}(u))$. 
Then $\Delta_{lex}(E)(u)=\Phi^{E}_{lex}(u) - \min(\Phi^{E}_{lex})$.
\end{Definition}
$\Delta_{lex}$ does not satisfy (Comm), but it satisfies (E1)-(E6), as well as (KP5)-(KP6). 
By exploiting the non-commutativity of $\Delta_{lex}$, both (Arb) and (Maj) can be phrased in a way to 
ensure that $\Delta_{lex}$ fails to satisfy them.

\section{Merging and infobases}
Our description of merging uses a representation of epistemic states as functions assigning
a plausibility ranking to the interpretations of $L$, but where do these plausibility rankings come
from? One way in which to generate them is by using the \emph{infobases}
of Meyer \cite{Meyer:99c}. An infobase is a finite list of wffs. Intuitively it is a structured
representation of the beliefs of an agent with a foundational flavour. It is assumed that 
every wff in an infobase is obtained independently. Meyer uses an
infobase to define a total preorder on $U$, which is then used to perform belief change.
However, we can also use an infobase to define an epistemic state. The idea is to consider
the number of times that an interpretation occurs as a model of one of the wffs in an infobase:
the more it occurs, the higher its plausibility ranking.
\begin{Definition}
For $u\in U$, define the $\mathit{IB}$-number $u_{\mathit{IB}}$ of $u$ as the number of elements $\alpha$ in an 
infobase $\mathit{IB}$ such that $\nvDash\alpha$ and $u\in M(\alpha)$, and let  
\[
\max(\mathit{IB})=\max\{u_{\mathit{IB}}\mid u\in U\}.
\] 
Now we define the epistemic state
associated with $\mathit{IB}$ as follows: for $u\in U, \Phi^{\mathit{IB}}(u)=\max(\mathit{IB}) - u_{\mathit{IB}}$.
\end{Definition}
Observe that the knowledge base associated with an epistemic state $\Phi^{\mathit{IB}}$ is always
consistent, regardless of whether the wffs in $\mathit{IB}$ are jointly consistent. We show that 
infobases seem to provide a natural setting in which to apply merging.

Firstly, define an \emph{infobase list} $\mathit{EB}=[\mathit{IB}_{1},\ldots,\mathit{IB}_{\left|\mathit{EB}\right|}]$ as a finite 
non-empty list of infobases and let $E^{\mathit{EB}}$ denote the epistemic list
$[\Phi^{\mathit{IB}_{1}},\ldots,\Phi^{\mathit{IB}_{\left|E\right|}}]$ of epistemic states
associated with the infobases occurring in $\mathit{EB}$. Then 
it can be verified that
$\Delta_{\Sigma}(E^{\mathit{EB}})=\Phi^{\mathit{IB}}$ where $\mathit{IB}=\bigsqcup_{i=1}^{\left|\mathit{EB}\right|}\mathit{IB}_{i}$. 

Secondly, Konieczny and Pino-P{\'e}rez \cite{Konieczny-ea:98a} give a convincing example to 
show that we may sometimes want to include, as models of $\delta(e)$, interpretations 
other than the models of the knowledge bases in $e$. Below is a scaled down version of their
example.
\begin{example}
We want to speculate on the stock exchange and we ask two equally reliable financial experts about 
two shares. Let the atom $p$ denote the fact that share 1 will rise and $q$ the fact that share 2
will rise. The first expert says that both shares will rise: $\phi_{1}=p\wedge q$, while the
second one believes that both shares will fall: $\phi_{2}=\lnot p\wedge\lnot q$. Intuitively it 
seems reasonable to conclude that both experts are right (and wrong) about exactly one
share, although we don't know which share in either case. That is, we require the result of the 
merging of these two knowledge bases to be such that
$M(\delta([\phi_{1}]\sqcup[\phi_{2}]))=\{10,01\}$.\footnote{%
We represent interpretations as sequences consisting of 0s (representing falsity) and 1s 
(representing truth), where the first digit in 
a sequence represents the truth value of $p$ and the second one the truth value of $q$.}
Observe that
$M(\delta([\phi_{1}]\sqcup[\phi_{2}]))\nsubseteq M(\phi_{1})\cup M(\phi_{2})$.
\end{example}
An analysis of this example shows that both experts are assumed to
make an implicit assumption of independence of the performance of the shares. Thus the
beliefs of the first expert is best expressed as the infobase $\mathit{IB}_{1}=[p,q]$ and the beliefs of 
the second expert as the infobase $\mathit{IB}_{2}=[\lnot p,\lnot q]$. The epistemic states obtained
from these two infobases are:
$\Phi^{\mathit{IB}_{1}}(11) = 0, \Phi^{\mathit{IB}_{1}}(10) = \Phi^{\mathit{IB}_{1}}(01) = 1, \Phi^{\mathit{IB}_{1}}(00) = 2$, and
$\Phi^{\mathit{IB}_{2}}(00) = 0, \Phi^{\mathit{IB}_{2}}(10) = \Phi^{\mathit{IB}_{2}}(01) = 1, \Phi^{\mathit{IB}_{2}}(11) = 2$. 
It can be verified that 
$\Delta_{\max}(E^{\mathit{EB}}) = \Delta_{Gmax}(E^{\mathit{EB}}) = \Delta_{R\Sigma}(E^{\mathit{EB}}) = \Phi$, 
where $\mathit{EB}=[\mathit{IB}_{1},\mathit{IB}_{2}]$, $\Phi(10) = \Phi(01) = 0$ and $\Phi(11) = \Phi(00) = 1$. So  
$\Delta_{R\Sigma}$, $\Delta_{\max}$ and $\Delta_{Gmax}$ yield the results corresponding to our 
intuition for this example. 
\section{Conclusion}
The merging operations we have constructed provide evidence that 
(E1)-(E4) may be regarded as basic postulates for merging operations on epistemic states. 
Furthermore, we regard (Arb) as an appropriate
postulate for the subclass of arbitration operations, (Maj) for the subclass of majority operations, 
and (Comm) for the subclass of commutative merging operations. 
The status of (E5) and (E6) is less clear. While all but one of the valid merging operations we have 
considered satisfy both, the fact that $\Delta_{R\Sigma}$ does not, suggests that they are not as
universally applicable as (E1)-(E4). Perhaps they should be seen as
picking out particular subclasses of merging operations in the way that (Arb), (Maj) and (Comm) do.
\bibliographystyle{aaai}
\bibliography{tmeyer}
\end{document}